\documentclass{article}

\usepackage{microtype}
\usepackage{graphicx}
\usepackage{booktabs} % for professional tables

\usepackage{amsmath,amsfonts,bm}

\def\eqref#1{equation~\ref{#1}}
\def\1{\bm{1}}

\DeclareMathAlphabet{\mathsfit}{\encodingdefault}{\sfdefault}{m}{sl}
\SetMathAlphabet{\mathsfit}{bold}{\encodingdefault}{\sfdefault}{bx}{n}

\usepackage{hyperref}
\usepackage{url}
\usepackage{booktabs}       % professional-quality tables
\usepackage{amsfonts}       % blackboard math symbols
\usepackage{nicefrac}       % compact symbols for 1/2, etc.
\usepackage{microtype}      % microtypography
\usepackage[inline]{enumitem}
\usepackage{subcaption}
\usepackage{graphicx}
\usepackage{wrapfig}
\usepackage{tabularx}
\usepackage{array}
\newcolumntype{Y}{>{\raggedright\arraybackslash}X} 
\usepackage{caption} 
\usepackage{xcolor}
\usepackage{hyperref}
\usepackage{siunitx}
\usepackage{multirow}
\usepackage{array, booktabs, makecell}
\usepackage{siunitx, mhchem}

\newlength\lengtha 

\usepackage{hyperref}
\usepackage{soul}

\usepackage[accepted]{icml2019}

\begin{document}

\twocolumn[
\icmltitle{Transfer Learning for Related Reinforcement Learning Tasks via Image-to-Image Translation}



\icmlsetsymbol{equal}{*}

\begin{icmlauthorlist}
\icmlauthor{Shani Gamrian}{bi}
\icmlauthor{Yoav Goldberg}{bi,goo}
\end{icmlauthorlist}

\icmlaffiliation{bi}{Computer Science Department, Bar-Ilan University, Ramat-Gan, Israel}
\icmlaffiliation{goo}{Allen Institute for Artificial Intelligence}

\icmlcorrespondingauthor{Shani Gamrian}{gamrianshani@gmail.com}
\icmlcorrespondingauthor{Yoav Goldberg}{yoav.goldberg@gmail.com}

\icmlkeywords{Machine Learning, ICML}

\vskip 0.3in
]



\printAffiliationsAndNotice{}  % leave blank if no need to mention equal contribution

\begin{abstract}
% Deep Reinforcement Learning has managed to achieve state-of-the-art results in learning control policies directly from raw pixels. However, despite its remarkable success, it fails to generalize, a fundamental component required in a stable Artificial Intelligence system. 
% Using the Atari game Breakout, we demonstrate the difficulty of a trained agent in adjusting to simple modifications in the raw image, ones that a human could adapt to trivially.
% In transfer learning, the goal is to use the knowledge gained from the source task to make the training of the target task faster and better. We show that using various forms of fine-tuning, a common method for transfer learning, is not effective for adapting to such small visual changes. In fact, it is often easier to re-train the agent from scratch than to fine-tune a trained agent. 
% We suggest that in some cases transfer learning can be improved by adding a dedicated component whose goal is to learn to visually map between the known domain and the new one.
% Concretely, we use Unaligned Generative Adversarial Networks (GANs) to create a mapping function to translate images in the target task to corresponding images in the source task. These mapping functions allow us to transform between various variations of the Breakout game, as well as between different levels of a Nintendo game, Road Fighter. We show that learning this mapping is substantially more efficient than re-training. A visualization of a trained agent playing Breakout and Road Fighter, with and without the GAN transfer, can be seen in \url{https://streamable.com/msgtm} and \url{https://streamable.com/5e2ka}.

Despite the remarkable success of Deep RL in learning control policies from raw pixels, the resulting models do not generalize. We demonstrate that a trained agent fails completely when facing small visual changes, and that fine-tuning---the common transfer learning paradigm---fails to adapt to these changes, to the extent that it is faster to re-train the model from scratch. 
We show that by separating the visual transfer task from the control policy we achieve substantially better sample efficiency and transfer behavior, allowing an agent trained on the source task to transfer well to the target tasks. The visual mapping from the target to the source domain is performed using unaligned GANs, resulting in a control policy that can be further improved using imitation learning from imperfect demonstrations. We demonstrate the approach on synthetic visual variants of the Breakout game, as well as on transfer between subsequent levels of Road Fighter, a Nintendo car-driving game. A visualization of our approach can be seen in \url{https://youtu.be/4mnkzYyXMn4} and \url{https://youtu.be/KCGTrQi6Ogo}.
\end{abstract}

\section{Introduction}

% \yg{consider dropping first paragraph, or shortening to one sentence. it does not read like a "serious" ML paper.}
% Transferring knowledge from previous occurrences to new circumstances is a fundamental human capability and is a major challenge for deep learning applications. A plausible requirement for artificial general intelligence is that a network trained on one task can reuse existing knowledge instead of learning from scratch for another task. For instance, consider the task of navigation during different hours of the day. A human that knows how to get from one point to another on daylight will quickly adjust itself to do the same task during night time, while for a machine learning system making a decision based on an input image it might be a harder task. That is because it is easier for us to make analogies between similar situations, especially in the things we see, as opposed to a robot that does not have this ability and its knowledge is based mainly on what it already saw. 

Deep reinforcement learning is becoming increasingly popular due to various recent success stories, such as the famous achievement of learning to play Atari 2600 video games from pixel-level input \citep{DBLP:journals/corr/MnihKSGAWR13}.
%Deep reinforcement learning has caught the attention of researchers in the past years for its remarkable success in achieving human-level performance in a wide variety of tasks. One of the field's famous achievements was on the Atari 2600 games where an agent was trained to play video games directly from the screen pixels and information received from the game \citep{DBLP:journals/corr/MnihKSGAWR13}. 
However, this approach depends on interacting with the environment a substantial number of times during training. Moreover, it struggles to generalize beyond its experience, the training process of a new task has to be performed from scratch even for a related one. Recent works have tried to overcome this inefficiency with different approaches such as, learning universal policies that can generalize between related tasks \citep{DBLP:journals/corr/SchaulQAS15}, as well as other transfer approaches \citep{DBLP:journals/corr/FernandoBBZHRPW17, DBLP:journals/corr/RusuRDSKKPH16}.

In this work, we first focus on the Atari game Breakout, in which the main concept is moving the paddle towards the ball in order to maximize the score of the game. We modify the game by introducing visual changes such as adding a rectangle in the middle of the image or diagonals in the background. From a human perspective, it appears that making visual changes that are not significant to the game's dynamics should not influence the score of the game. 
%a player who mastered the original game should be able to trivially adapt to such visual variants. 
We show that the agent fails to transfer. Furthermore, fine-tuning, the main transfer learning method used today in neural networks, also fails to adapt to the small visual change: the information learned in the source task does not benefit the learning process of the very related target task, and can even decelerate it. %The algorithm behaves as if these are entirely new tasks.

Our second focus is attempting to transfer agent behavior across different levels of a video game: can an agent trained on the first level of a game use this knowledge and perform adequately on subsequent levels? We explore the Nintendo game Road Fighter, a car racing game where the goal is to finish the track before the time runs out without crashing. The levels all share the same dynamics, but differ from each other visually and in aspects such as road width. Similar to the Breakout results, an agent trained to play the first level fails to correctly adapt its past experience, causing the learned policy to completely fail on the new levels. 

To address the generalization problem, we propose 
to isolate the visual component and perform zero-shot analogy-based transfer. 
%zero-shot transfer.
Concretely, the agent %learns to 
transfers between the tasks by learning to visually map images from the target task back to familiar corresponding images from the source task. Such mapping is naturally achieved using Generative Adversarial Networks (GANs) \citep{NIPS2014_5423}. %, one of the most popular methods for the image-to-image translation that is being used in various computer vision tasks} \citep{DBLP:journals/corr/ZhuPIE17, DBLP:journals/corr/ZhouXYFHH17, volz:gecco2018}.
%such as style transfer \citep{DBLP:journals/corr/ZhuPIE17, DBLP:journals/corr/KimCKLK17}, object transfiguration \citep{DBLP:journals/corr/ZhouXYFHH17}, photo enhancement \citep{DBLP:journals/corr/LedigTHCATTWS16} and more recently, video game level generation \citep{volz:gecco2018}. 
  In our setup, it is not realistic to assume paired images in both domains, calling for the use of Unaligned GANs \citep{DBLP:journals/corr/LiuBK17, DBLP:journals/corr/ZhuPIE17, DBLP:journals/corr/KimCKLK17, DBLP:journals/corr/YiZTG17}.

The approach allows the agent to effectively apply its source domain knowledge in the target domain without additional training. For cases where the visual analogy is insufficient for performing optimally, or where the GAN fails to produce sufficiently accurate mappings, we treat the resulting policy as an imperfect demonstration and further improve it using an imitation learning algorithm tailored to the imperfect demonstration scenario. The code is available at \url{https://github.com/ShaniGam/RL-GAN}.
%Using this approach we manage to transfer between similar tasks with no additional learning.
%We further introduce an imitation learning from imperfect demonstration algorithm to improve the results on tasks that are visually difficult to map.

\textbf{Contributions} This work presents three main contributions. First, in Section \ref{DRL}, we demonstrate how an agent trained with deep reinforcement learning algorithms fails to adapt to small visual changes, and that the common transfer method of fine-tuning fails as well. Second, in Section \ref{GAN}, we propose to separate the visual mapping from the game dynamics, resulting in a new transfer learning approach for related tasks based on visual input mapping. Third, in Section \ref{IL_A2C}, we suggest an imitation learning from imperfect demonstrations algorithm for improving the visual-transfer-based policy in a sample efficient manner. %  approach for collecting and using imperfect demonstration for acceleration and better performances of the agents.
We evaluate this approach on Breakout and Road Fighter in Section \ref{GAN_exp}, and present the results comparing to different baselines. We show that our visual transfer approach is much more sample efficient than the alternatives. Moreover, we use our method as an evaluation setup for unaligned GAN architectures, based on their achieved performance on concrete down-stream tasks.
%Third, in section \ref{GANEval}, we suggest an evaluation setup for unaligned GAN architectures, based on their achieved performance on concrete down-stream tasks.

\begin{figure}
\centering
\subcaptionbox{\label{fig:1a}}{\includegraphics[width=0.75in, height=2.0cm]{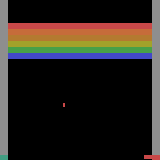}}\hspace{0.7em}%\vspace{0.3em}
\subcaptionbox{\label{fig:1b}}{\includegraphics[width=0.75in, height=2.0cm]{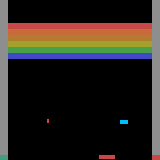}}\hspace{0.7em}%\vspace{0.1em}
\subcaptionbox{\label{fig:1c}}{\includegraphics[width=0.75in, height=2.0cm]{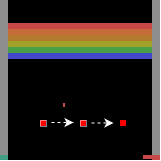}}\\%\vspace{0.3em}
\subcaptionbox{\label{fig:1d}}{\includegraphics[width=0.75in, height=2.0cm]{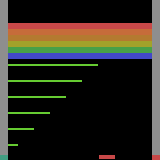}}\hspace{0.7em}
\subcaptionbox{\label{fig:1e}}{\includegraphics[width=0.75in, height=2.0cm]{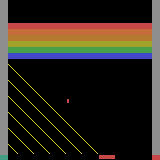}}
\caption{Various variations of the Breakout game: (a) Standard version, (b) A Constant Rectangle - a rectangle in the same size as the bricks is added to the background in a predefined location, (c) A Moving Square - a square is added to the background and its location changes to one of three predefined locations every $1000$ steps, (d) Green Lines - green lines in different sizes are drawn in the background, (e) Diagonals - diagonals are drawn in the left side of the background.}
\label{figure1}
\vspace{-2em}
\end{figure}

\section{Generalization failures of Deep RL}
\label{DRL}

Many Breakout variations that involve the same dynamics can be constructed. The main idea is to make modifications that are not critical for a human playing the game but are for the algorithm that relies on visual inputs. We demonstrate the difficulty of deep reinforcement learning to generalize using 4 types of modifications as presented in Figure \ref{figure1}.

% \subsection{Setup}
% \label{A3Csetup}

% For all the experiments in this section forward we use the Asynchronous Advantage Actor-Critic (A3C) algorithm \citep{DBLP:journals/corr/MnihBMGLHSK16}. 
% %taking advantage of being faster than Deep Q-Network (DQN) \citep{DBLP:journals/corr/MnihKSGAWR13}. 
% The A3C learns the policy and the state-value function using parallel actor-learners exploring different policies for the acceleration and stability of the training.

% We rescale the image to $80 \times 80$ and keep the RGB colors for more realistic images. We use $32$ actor learners, a discount rate of $0.99$, learning rate of $0.0001$, $20$-step returns, and entropy regularization weight of $0.01$. The A3C variation we choose is the LSTM-A3C network. We use the standard high-performance architecture implemented in \citep{pytorchaaac}.
%: $4$ convolutional layers with $32$ $3x3$ filters and a stride length of $2$, connected to an LSTM layer with $256$ cells whose output is linearly fed into two output layers, one produces the next optimal action (actor) and the other, the approximated value (critic). 

%\subsection{Transfer-Learning via Fine-Tuning}
\textbf{Transfer-Learning via Fine-Tuning.} For all the experiments in this section forward we use the Asynchronous Advantage Actor-Critic (A3C) algorithm \citep{DBLP:journals/corr/MnihBMGLHSK16}. The A3C learns the policy and the state-value function using parallel actor-learners exploring different policies for the acceleration and stability of the training. Details of the setup can be seen in the appendix (\ref{A3C-setup}). This setup successfully trains on Breakout, reaching a score of over $400$ points. However, when a network trained on the original game is presented with the game variants, it fails completely, reaching to a maximum score of only $3$ points. This shows that the network does not necessarily learn the game's concepts and heavily relies on the images it receives.

The common approach for transferring knowledge across tasks is fine-tuning. We experiment with common techniques used in deep learning models. In each setting, we have a combination of frozen and fine-tuned layers (Partial/Full) as well as layers that are initialized with the target's parameters and layers that are initialized with random values (Random). Our settings are inspired by \citep{Yosinski:2014:TFD:2969033.2969197}. We train each one of the tasks (before and after the transformation) for $60$ million frames, and our evaluation metric is the total reward the agents collect in an episode averaged by the number of episodes, where an episode ends when the game is terminated or when a number of maximum steps is reached. We periodically compute the average during training. Details are available in the appendix (\ref{ft-settings}).
% We consider the following settings:
% \begin{itemize}
%   \item From-Scratch: The game is being trained from scratch on the target game.
%   \item Full-FT: All of the layers are initialized with the weights of the source task and are fine-tuned on the target task.
%   \item Random-Output: The convolutional layers and the LSTM layer are initialized with the weights of the source task and are fine-tuned on the target task. The output layers are initialized randomly.
%   \item Partial-FT: All of the layers are initialized with the weights of the source task. The three first convolutional layers are kept frozen, and the rest are fine-tuned on the target task.
% \item Partial-Random-FT: The three first convolutional layers are initialized with the weights of the source task and are kept frozen, and the rest are initialized randomly.
% \end{itemize}

% \subsection{Results}
%\setlength\belowcaptionskip{-2ex}
%\vspace{-6em}

\begin{figure}[t!]
\centering
\subcaptionbox{A Constant Rectangle\label{fig:2a}}{\includegraphics[width=1.37in]{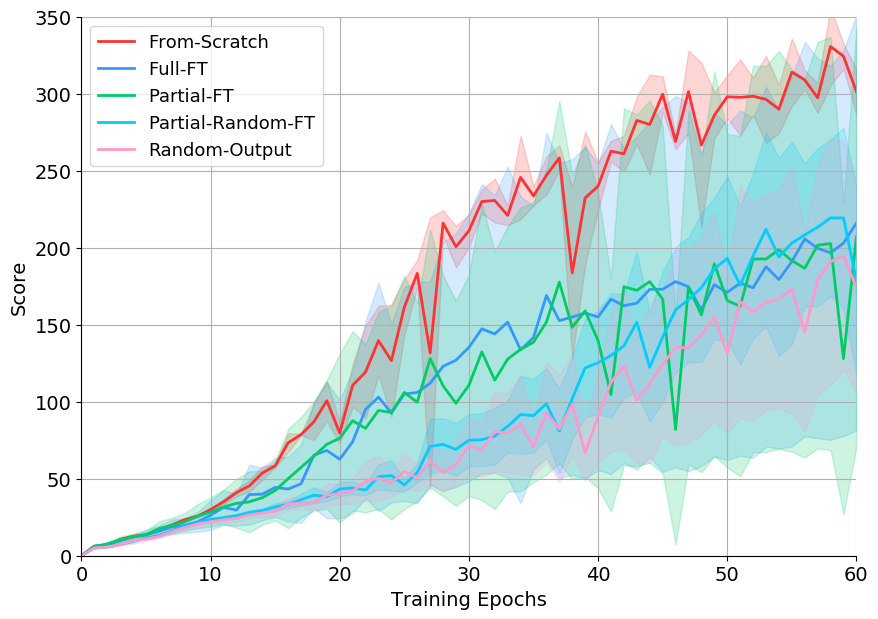}}\hspace{0.1em}
\vspace{0.3em}
\subcaptionbox{A Moving Square\label{fig:2b}}{\includegraphics[width=1.37in]{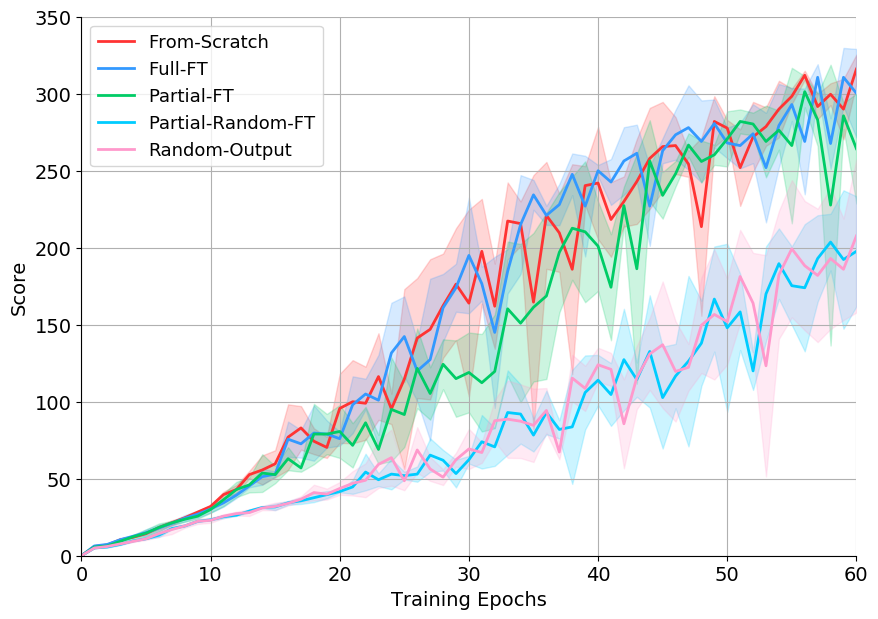}}\hspace{0.1em}
\vspace{0.3em}
\subcaptionbox{Green Lines\label{fig:2c}}{\includegraphics[width=1.37in]{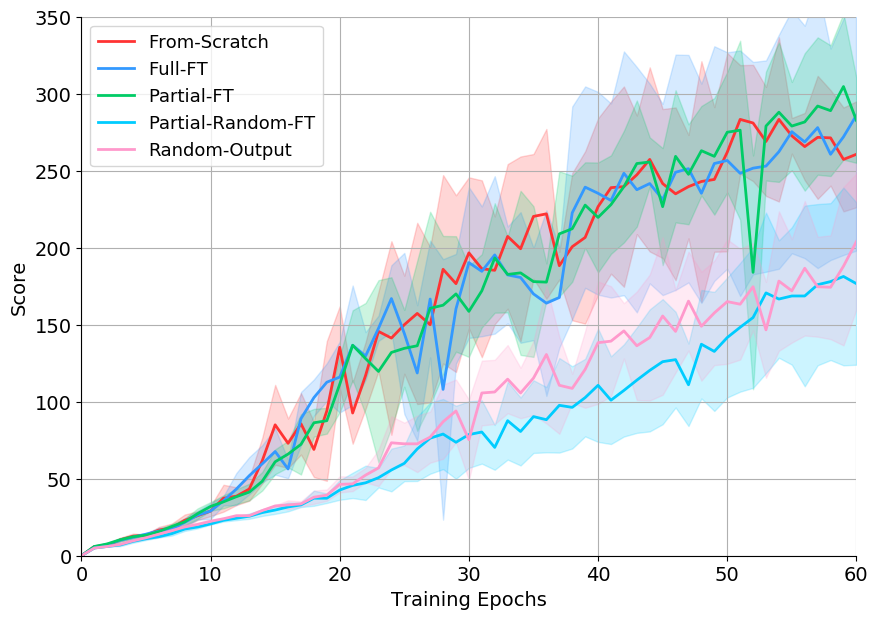}}\hspace{0.1em}
\subcaptionbox{Diagonals\label{fig:2d}}{\includegraphics[width=1.37in]{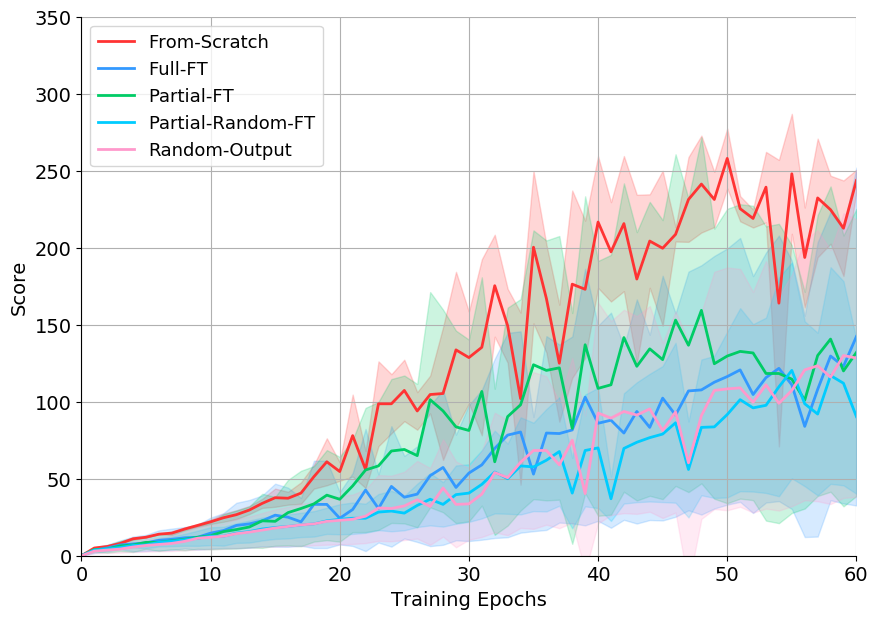}}
\caption{A comparison between the different transfer techniques on Breakout. The y-axis shows the average reward per episode of Breakout during training. The x-axis shows the total number of training epochs where an epoch corresponds to 1 million frames. The plots are averaged on 3 runs with different random seeds. Each curve is the average and its background is the standard deviation.}
\label{figure2}
\vspace{-1em}
\end{figure}

\textbf{Results.} The results presented in Figure \ref{figure2} show a complete failure of all the fine-tuning approaches to transfer to the target tasks. In the best scenarios the transfer takes just as many epochs as training from scratch, while in other cases starting from a trained network makes it \emph{harder} for the network to learn the target task. As the graphs show, some of the modification interfere more than others. For example, Figure \ref{fig:2a} shows that adding a simple rectangle can be destructive for a trained agent: while training from scratch consistently and reliably achieves scores over $300$, the settings starting from a trained agent struggle to pass the $200$ points mark within the same number of iterations, and have a very high variance. We noticed that during training the agent learns a strategy to maximize the score with a minimum number of actions. 
None of the experiments we performed showed better results when the layers in the network were fine-tuned, and some showed negative transfer which is a clear indication of an overfitting problem. The A3C model learned the detail and noise in the training data to the extent that it negatively impacted the performance of the model on new data. Our results and conclusions drawn from them are consistent with the results shown when a similar approach was used on Pong \citep{DBLP:journals/corr/RusuRDSKKPH16}. In addition to Breakout/A3C, we also attempted to transfer between a model trained with the synchronous actor-critic variant, A2C, from the first to advanced level of Road Fighter, where the backgrounds change but the dynamics remains the same. This resulted with $0$ points on each of the levels, a complete failure of the agent to re-use the driving techniques learned on the first levels on following ones.

\section{Analogy-based Zero-Shot Transfer} 
\label{GAN}

An agent capable of performing a task in a source domain is now presented with a new domain. Fine-tuning the agent on the target domain fails to transfer knowledge from the source domain. We propose to separate the visual transfer from the dynamics transfer.
To perform well, the agent can try and make analogies from the new domain to the old one: after observing a set of states (images) in the new domain, the agent can learn to map them to similar, familiar states from the source domain, and act according to its source domain policy on the mapped state. 

More concretely, given a trained policy $\pi(a|s;\theta)$ with trained parameters $\theta$ proposing an action $a$ for source domain states $s\in\mathcal{S}$, we wish to learn a mapping function $G : \mathcal{T} \mapsto \mathcal{S}$ from target domain states $t \in \mathcal{T}$ such that interacting with the environment $\mathcal{T}$ by applying the policy $\pi(a|G(t);\theta)$ will result in a good distribution of actions for the states $\mathcal{T}$, as indicated by high overall scores.
In other words, we seek a mapping function $G$ that allows us to re-use the same policy $\pi_\theta$ learned for source environment $\mathcal{S}$ when interacting with the target environment $\mathcal{T}$.

As both the source and target domain items are images, we heuristically learn the function $G$ by collecting sets of images from $\mathcal{S}$ and $\mathcal{T}$ and learning to visually map between them using Unaligned GAN \citep{DBLP:journals/corr/LiuBK17, DBLP:journals/corr/ZhuPIE17, DBLP:journals/corr/KimCKLK17, DBLP:journals/corr/YiZTG17}. We use the scores obtained from interacting with the environment via $\pi(a|G(t);\theta)$ for the GAN model selection and stopping criteria.
\vspace{-1em}
\paragraph{Unsupervised Image-to-image Translation}
In this work, we focus on learning setups that receive only raw image data, without additional domain knowledge about objects or game dynamics. This prohibits us from using supervised paired GANs \cite{DBLP:journals/corr/IsolaZZE16} for learning the mapping function $G$: we cannot collect the needed supervision of corresponding $(s,t)$ pairs. Instead, we use unaligned GANs \citep{DBLP:journals/corr/ZhuPIE17, DBLP:journals/corr/LiuBK17, DBLP:journals/corr/KimCKLK17, DBLP:journals/corr/YiZTG17}, in which the learner observes two sets of images, one from each domain, with the goal of learning to translate images in one domain to images in another.

All major approaches to the unaligned image-to-image translation use the Cycle-Consistency principle.
We have two mapping (encoding) functions $G_1: T \mapsto S$ and $G_2: S \mapsto T$ where $S=\{s_i\}_{i=1}^N$ is a set of images collected from the source task and $T=\{t_j\}_{j=1}^M$ is a set of images collected from the target task. The goal is to generate an image $s'$, for any given $t \in T$ where $G_1(t)=s'$, that is indistinguishable from $s \in S$. The cycle consistency principle relies on the assumption that the two functions, $G_1$ and $G_2$ are inverses of each other. It encourages unsupervised mapping by forcing $G_2(G_1(t))=t$ and $G_1(G_2(s))=s$ where $s$ and $t$ are the input images. The second component of the GAN architecture are the discriminators $D_1$ and $D_2$ aiming to distinguish between images generated by $G_1$ and $G_2$ and the real images from the target and source distributions respectively.

In the following experiments, we use the UNIT framework \citep{DBLP:journals/corr/LiuBK17}, which we found to perform well for the Breakout tasks (in section \ref{GANEval} we explicitly compare the UNIT and CycleGAN approaches on both the Breakout and Road Fighter transfer tasks). 
% \yg{I think can delete from here to end of paragraph.}A distinguishing element in the UNIT framework is the shared-latent space assumption, according to which there is a shared-latent space consisting a shared latent code $z$ for any pair of images $s$ and $t$ that can be recovered from this code. This shared-latent space is represented as the weights of the last few layers of the encoding network and the few first layers of the decoding networks, and is learned by using Variational Autoencoders (VAEs). This sharing strongly ties the images in the source and target domain to each other, encouraging mappings that preserve similarities across domains. In contrast, the CycleGAN architecture \citep{DBLP:journals/corr/ZhuPIE17} does not make the shared space assumption and instead the generators are trained independently with two separate networks.
% For further information of the unaligned GAN architectures, see the original papers.

\paragraph{GAN Training}

%\textbf{Datasets.} %To train the GAN we need a group of images from the source domain and a group of images from the target domains. Since assembling two groups of matching images can be difficult and even impossible, we use an unsupervised GAN. 
The Unaligned GAN training dataset requires images from both domains. We collect images from the source domain by running an untrained agent and collecting the observed images, and we do similarly for the target domain.
% The number of collected images should balance between two objectives:
% on the one hand, we want to minimize the interaction with the environment (take a small number of images), and on the other hand, it is essential for us to have a diverse dataset. \sgp{We accomplish that by selecting a number of frames that allows the agent to reach a few game point.} 
% \yg{maybe delete all the "The number of collected images should balance..." part until here altogether? why is this important?}
We repeat this procedure for every target task, and create a source-target dataset for each.

%\textbf{Setup.} 
For our experiments we use the same architecture and hyper-parameters proposed in the UNIT paper. We initialize the weights with Xavier initialization \citep{Glorot10understandingthe}, set the batch size to $1$ and train the network for a different number of iterations on each task.
%Some tasks are harder than others, the more changes exist in the frames the harder it is for the GAN to learn the mapping between the domains. 
% \yg{From here on it is repetitive with the next part, can remove.} However, our evaluation metric, testing the agent with the generated images, is a clear indication of how hard each task is and the number of iterations needed is based on the results of this evaluation.

\textbf{Concrete GAN Evaluation Criterion.} %We use GAN training to learn a mapping function $G$.
GAN training, and unaligned GANs training in particular, are unstable and it is challenging to find a good loss-based stopping criteria for them.  %To find the best GAN architecture as well as the number of training iterations needed for each task, we need to evaluate the model on each iteration. 
A major issue with GANs is the lack of an evaluation metric that works well for all models and architectures, and which can assist in model selection. % Different works use different methods that were suitable for their types of data.
Fortunately, our setup suggests a natural evaluation criteria: we run the source agent without any further training while using the model to translate each image of the target task back to the source task and collect the rewards the agent receives during the game when presented with the translated image. We use the total accumulated rewards (the score) the agent collects during the game as the criteria for the GAN's model quality, for model selection. In section \ref{GANEval} we use this criteria to compare unaligned GAN variants.

\section{Imitation Learning}
\label{IL_A2C}

The visual-analogy transfer method allows the agent to re-use its knowledge from the source domain to act in the target domain, resulting in adequate policies.
%Using our method, we manage to transfer between related tasks by mapping every input image from target to source. While our approach works, it is limited to the GAN generation ability, which might set the performances of the agent to be below training from scratch. 
However, it is limited by the imperfect GAN generation and generalization abilities, which, for more challenging visual mappings, may result in sub-optimal policies.
%We propose to improve the transfer process of the agent by collecting a small amount of demonstration and using them to accelerate the learning process on the target tasks.

% \ygp{We propose to use the visual-transfer based policy as an \textit{imperfect demonstration}, and use imitation learning from this policy to further improve the agent, while maintaining better sample efficiency than learning from scratch.}

We propose to use the visual-transfer based policy to create \textit{imperfect demonstrations}, and use imitation learning from these demonstrations to further improve the agent, while maintaining better sample efficiency than learning from scratch.

\begin{algorithm}[t]
\caption{Imitation Learning}\label{alg:IL}
\begin{algorithmic}
\STATE \textbf{Input:} a source trained network $\hat{\theta}$, a generator trained with GANs $Gen$
\STATE Initialize replay buffer $\mathcal{D}\gets \emptyset$, trajectory buffer $\mathcal{T}\gets \emptyset$
\textbf{\LineComment{Collecting trajectories}}
\FOR{$i=1$ to $Trajectories$}
    \STATE Get initial state $s_0$
    \REPEAT
        %\STATE $s_{gen}\gets Gen(s_t)$
        \STATE Execute an action $a_t,r_t,s_{t+1} \sim \pi_{\hat{\theta}}(a_t|Gen(s_{t}))$
        \STATE Store transition $\mathcal{T}\gets \mathcal{T} \cup {(s_t,a_t,r_t)}$
        \STATE $s_t\gets s_{t+1}$
    \UNTIL{$s_t$ is terminal}
    \IF{$r_t > \beta_1 R_\mathcal{T}$}
    \STATE Compute returns $R_t=\sum_{k}^{\infty}\gamma^{k-t}r_k$
    \STATE $\mathcal{D}\gets \mathcal{D} \cup (s_t,a_t,R_t)$ for all $t$ in $\mathcal{T}$
    \ENDIF
    \STATE Clear trajectory buffer $\mathcal{T}\gets \emptyset$
\ENDFOR
\textbf{\LineComment{Supervised Training}}
\STATE Initialize target network weights $\theta$ randomly
\FOR{$i=1$ to $Supervised\_Iterations$}
    \STATE Train on $\mathcal{D}$ with $\mathcal{L}_{IL}$ using SGD and batch size $b$
\ENDFOR
\textbf{\LineComment{RL with A2C}}
\FOR{$e=1$ to $Epochs$}
    \FOR{$t=1$ to $Steps$}
        \STATE Execute an action $a_t,r_t,s_{t+1} \sim \pi_{\theta}(a_t|s_t)$
        \STATE $s_t\gets s_{t+1}$
    \ENDFOR
%    \STATE Compute returns $R_t=\sum_{k}^{\infty}\gamma^{k-t}r_k$
    \textbf{\LineComment{On-policy updates}}
    \STATE Update $\theta$ according to $\mathcal{L}_{a2c}$ using RMSprop
    \textbf{\LineComment{Off-policy updates}}
    \IF{$t\mod op\_interval = 0$ AND $\hat{R} < \beta_2 R_\mathcal{T}$}
        \STATE Train on $\mathcal{D}$ with $\mathcal{L}_{IL}$ using SGD and batch size $b$
    \ENDIF
\ENDFOR
\end{algorithmic}
\end{algorithm}

\textbf{Accelerating RL with Imitation Learning.} We combine RL training with Imitation Learning similarly to \citep{DBLP:journals/corr/HesterVPLSPSDOA17, DBLP:journals/corr/abs-1802-05313, DBLP:journals/corr/abs-1806-05635}, but with special considerations for the imperfect demonstration scenario. More specifically, we focus on combining an actor-critic approach with supervised training on imperfect demonstrations. Our process (Algorithm \ref{alg:IL}) consists of 3 stages. We start by collecting trajectories of the transferred agent interacting with the environment of the target task. The trajectories are collected while following a stochastic policy, to improve coverage. Each trajectory stochastically follows a visually-transferred agent for an entire game. We find that a small number of 5 trajectories is sufficient for obtaining strong results. For each trajectory, in each time step $t$, we collect the state of the target task $s_t$, action $a_t$, and real values $R_t$ as a triple $(s_t, a_t, R_t)$ and store it in a buffer $\mathcal{D}$. We use $\mathcal{D}$ to train a new agent on the target task by imitating the action and value of each observation from the demonstrations using supervised learning. After a limited number of iterations we switch to RL training, while combining on-policy A2C updates and off-policy supervised updates from the buffer. To account for the imperfect demonstrations, we stop the off-policy updates and move to exclusive on-policy RL training once the agent perform better than the demonstrations.

%We train the model using two different approaches and optimize a different objective functions in each. 
The objective function of the off-policy updates before and during the RL training is given by:\\
$\mathcal{L}_{IL} = \mathbb{E}_{s,a,R\sim\mathcal{D}}[\mathcal{L}_{IL_{policy}} + \frac{1}{2}\mathcal{L}_{IL_{value}}]$\\
%$\hat{a} = \pi_\theta(s)$ \textcolor{mygray}{//$a$ is Multi-Binary}
$\mathcal{L}_{IL_{policy}} = \frac{1}{|a|}\sum_{k=0}^{|a|}a_k\log{(\hat{a}_k)}+(1-a_k)\log{(1-\hat{a}_k)}$ \\
$\mathcal{L}_{IL_{value}} = {(R-V_\theta(s))}^2$\\
where $\hat{a} = max_a\pi_\theta(a|s)$, $R_t=\sum_{k}^{\infty}\gamma^{k-t}r_k$ and $\pi_\theta(a|s)$, $V_\theta(s)$ are the policy and value functions parameterized by $\theta$.
For the on-policy updates we use the A2C loss which is given by:\\
$\mathcal{L}_{a2c} = \mathbb{E}_{s_t,a_t\sim\pi_{\theta}}[\mathcal{L}_{a2c_{policy}} + \frac{1}{2}\mathcal{L}_{a2c_{value}}]$\\
$\mathcal{L}_{a2c_{policy}} = -\log{\pi_{\theta}(a_t|s_t)}(V_t^n-V_\theta(s_t)) - \alpha\mathcal{H}_t^{\pi_\theta}$\\
$\mathcal{L}_{a2c_{value}} = {(V_t^n-V_\theta(s_t))}^2$ \\
where $V_t^n = \sum_{d=0}^{n-1}\gamma^dr_{t+d}+\gamma^nV_\theta(s_{t+n})$ is the n-step bootstrapped value. $\mathcal{H}_t^{\pi_\theta} = -\sum_a\pi(a|s_t)\log\pi(a|s_t)$ is the entropy and $\alpha$ is the entropy regularization factor.

\begin{table*}[t!]
\caption{The score and number of frames needed for it of: the source task (Source), target task when initialized with the source task' network parameters with no additional training (Target) and the target task when initialized with the source task' network parameters where every frame is translated to a frame from the source task (Target with GANs).}\label{table:mr}
\begin{tabularx}{\textwidth}{YYlYYYY@{}}
\toprule
\multicolumn{2}{c}{\bfseries Source}
%&\textbf{Target Task} 
&\multicolumn{1}{c}{\bfseries Target Task}
&\multicolumn{2}{c}{\bfseries Target}
&\multicolumn{2}{c}{\bfseries Target with GANs} \\
\cmidrule(l){1-2} \cmidrule(l){3-3} \cmidrule(l){4-5} \cmidrule(l){6-7}
Frames 
&Score
&
&Frames 
&Score
&GAN iterations
&Score
\\
\midrule
43M & 302 & A Constant Rectangle & 0 & 3 & \textbf{260K} & \textbf{362} \\
43M & 302 & A Moving Square & 0 & 0 &  \textbf{384K} & \textbf{300} \\
43M & 302 & Green Lines & 0 & 2 & \textbf{288K} & \textbf{300} \\
43M & 302 & Diagonals & 0 & 0 & \textbf{380K} & \textbf{338} \\
\bottomrule
\end{tabularx}
\vspace{-1em}
\end{table*}

\begin{figure}[!ht]
\centering
\subcaptionbox{\label{fig3:a}}{\includegraphics[width=0.7in]{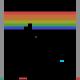}
\includegraphics[width=0.7in]{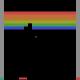}}\hspace{.65em}
\subcaptionbox{\label{fig3:b}}{\includegraphics[width=0.7in]{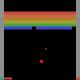}
\includegraphics[width=0.7in]{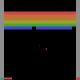}}\hspace{.65em}
\subcaptionbox{\label{fig3:c}}{\includegraphics[width=0.7in]{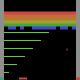}
\includegraphics[width=0.7in]{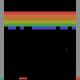}}\hspace{.65em}
\subcaptionbox{\label{fig3:d}}{\includegraphics[width=0.7in]{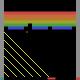}
\includegraphics[width=0.7in]{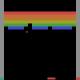}}\hspace{.65em}
\subcaptionbox{\label{fig3:e}}{\includegraphics[width=0.7in]{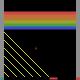}
\includegraphics[width=0.7in]{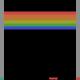}}\hspace{.65em}
\subcaptionbox{\label{fig3:f}}{\includegraphics[width=0.7in]{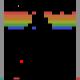}
\includegraphics[width=0.7in]{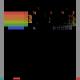}}
\caption{Illustration of a frame taken from the target task (left) and its matching frame of the source task generated with GANs (right) for each one of the Breakout variations. (a)-(d) demonstrate successes, while (e) and (f) show failure modes of the unaligned GAN. In (e) the ball in the input image is not generated in the output and in (f) not all bricks are generated, and some of the generated bricks appear smudged.}
\vspace{-1em}
\end{figure}

\textbf{Imperfect Demonstrations.} While learning from expert demonstrations is a popular technique for learning from good behaviors \cite{DBLP:journals/corr/abs-1011-0686, DBLP:journals/corr/HoE16, DBLP:journals/corr/abs-1805-01954}, learning from imperfect demonstrations remains a challenging task. We design the algorithm to ensure that the demonstrations benefit and not harm the learning process. First, 
we collect trajectories by sampling actions rather than following the best ones, for achieving diversity.
%we collect trajectories by letting the agent sample the actions from a distribution for diversity.
Stochastic policy leads to more exploration but can also leads to bad trajectories with low scores. %We pick the trajectories with scores higher than 
We discard trajectories with scores lower than
$\beta_1 R_\mathcal{T}$ ($\beta_1=0.75$), where $R_\mathcal{T}$ is the score of a trajectory collected with a deterministic policy (according to Table \ref{table:RF}) and $\beta_1$ determines how much below the maximum is considered a good trajectory. Second, we combine on-policy and off-policy updates for acceleration and stabilization of the training. 
Imitating the demonstration behavior is useful only if it is better than the agent's current policy. Additionally, to allow substantial improvements over the demonstration the agent may need to deviate from it into other policies that might be worse in order to eventually find a better one. Combining these rationales, we limit the off-policy updates to cases in which the mean reward of all agents 
$\hat{R}$ is smaller than the $\beta_2 R_\mathcal{T}$ ($\beta_2=0.6$).\footnote{The off-policies may turn on again if the policy degrades to below the demonstration-based reward.}

%Imitating a behaviour from demonstrations 
%is beneficial as long as its better than the agent's current behaviour. For this reason, we limit the off-policy updates to be only when the mean reward of all agents $\hat{R}$ is smaller than the $\beta_2 R_\mathcal{T}$.\footnote{The off-policies may turn on again if the policy degrades to below the demonstration-based reward.}

\section{Experiments}

\label{GAN_exp}
We apply the approach to the Breakout variants (where the visual transfer was sufficient to achieve perfect policy, despite deficiencies in GAN behavior) and to the much more challenging task of transfer between levels of the Road Fighter game, where the visual transfer resulted in an adequate agent, which could then be improved substantially by the imitation learning.
%We examine how well the agent does when receiving translated frames generated by the generator trained with GANs. Since we initialized the layers with the values of the trained network, we assume that the success of the agent is dependent on the similarity between the generated and the source task's frames. We first test our approach on Breakout, evaluating its ability to remove the changes added in the images. Second, we  challenge our method even more on Road Fighter, where the goal is to transfer between different environments. 

\subsection{Breakout}

The visual transfer goal in the Breakout variants is removing the visual modifications in each of the target games and mapping back to the unmodified source game.  %Although the variations share many similarities, 
Some tasks turned out to be more challenging to the GAN than others: the Green Lines variation hides parts of the ball in some frames making the training harder, while the Rectangle variation required less training since the number of changed pixels in the image is small. 

Overall, the translation tasks---despite their apparent simplicity---proved to be surprisingly challenging for the GANs.
While the input and output spaces are highly structured, the network does not have any information about this structure. Instead of learning a ``leave game objects intact'' policy, it struggles to model and generate them as well. The most common problem was the generator adding or removing blocks from the original image, and in general failing to correctly model block structure (see Fig. \ref{fig3:f} and the video). Another issue was with correctly tracking the ball location (Fig. \ref{fig3:e}).\footnote{We believe that this highly structured setting exposes an inherent deficiency in unaligned GAN models, suggesting an interesting avenue for future GAN research.} 
%During testing, we encountered problems with the images generation that we did not observe during the GAN training. The translation task we attempted to perform was supposedly simple -- search for the differences between the domains shared by all images, change them to the way they are in the opposite domain and leave everything else the same. Unfortunately, since the network does not have any prior information about objects in the image, it struggles to generate them even if they were not changed. The most common problem we had was that the generator generated bricks that were supposed to be removed from the game, and in some cases, they were very noisy in the generated image (Fig. \ref{fig3:f}). Another problem was the ball generation and more specifically, the location of the generated ball. Since the ball is small and changes its position often, it was hard for the generator, trained with unaligned pairs, to decide if and where to locate it in the generated image (Fig. \ref{fig3:e}).
%These issues and others eventually caused the agent to fail in following the policies it learned on the source task. We found that more training leads to better results for some of the variations and so the number of iterations needed was different for each variation.
Despite these limitations, the learned visual mapping was sufficient for the trained Breakout agent to achieve high scores.

Table \ref{table:mr} shows the results of a test game played by the agent with and without the GAN transfer. 
The source agent was trained until it reached 300 points, which we consider to be a high score. This required 43M frame interactions. When applied to the target tasks, the agent fails with scores $\leq$ 3. As discussed in Section \ref{DRL}, training from scratch for these tasks will require a similar number of frames. With the GAN based transfer the agent achieves scores $\geq$ 300, while observing only $100k$ target task frames and performing hundreds of thousands of GAN iterations, a 100x fold increase in sample efficiency.
%We stop the training after reaching $300$ points, which we consider to be a high score. As the results show, the source game trained from scratch requires ten of millions of images to achieve such score comparing to the target task trained with GANs that only needs a few hundreds of thousands---a 100x fold increase in sample efficiency. Moreover, the frames the GAN was trained on were limited to the first games in which the A3C network was not trained, and yet it managed to generalize to more advanced stages of the game.

\subsection{Road Fighter}

\begin{figure}[h!]
\centering
\subcaptionbox*{}{\includegraphics[width=0.71in, height=2.0cm]{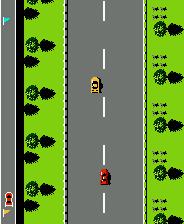}}\hspace{0.3em}
\subcaptionbox*{}{\includegraphics[width=0.71in, height=2.0cm]{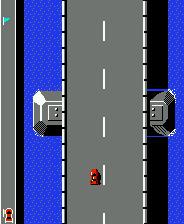}}\hspace{0.3em}
\subcaptionbox*{}{\includegraphics[width=0.71in, height=2.0cm]{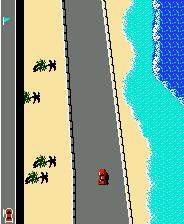}}\hspace{0.3em}
\subcaptionbox*{}{\includegraphics[width=0.71in, height=2.0cm]{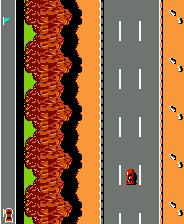}}
\caption{Road Fighter levels from left to right: Level 1, Level 2, Level 3 and Level 4.}
\label{figure4}
\end{figure}

While the Breakout variants work well to demonstrate the RL transfer failure cases, we consider them as ``toy examples''. We proceed to demonstrate the effectiveness of our transfer method on a ``real'' task: learning to transfer between different levels of the Road Fighter game. Road Fighter contains 4 different levels (Fig. \ref{figure4}), each with a different background where some are more difficult than others. The levels mostly differ visually and all have the same rules and require the same driving techniques. %Thus, we believe that these techniques should sustain when playing a new level. 
We train an RL agent to play the first level of the game. To master the game, the agent has to acquire 3 main capabilities: driving fast, avoiding collision with obstacles, and if a car collision occurs reacting fast to avoid crashing. We use the A2C algorithm, the synchronous version of the Advantage Actor-Critic which performs better on this game than A3C, reaching over $10,000$ game points on level 1. Training an RL agent to play this level requires observing over 100M frames.
The trained agent fails completely on more advanced levels, reaching a score of 0.

\begin{figure}[ht!]
\centering
\subcaptionbox{\label{fig:5a}}{\includegraphics[width=0.7in]{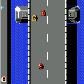}
\includegraphics[width=0.7in]{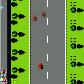}}\hspace{.65em}
\subcaptionbox{\label{fig:5b}}{\includegraphics[width=0.7in]{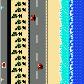}
\includegraphics[width=0.7in]{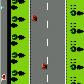}}\hspace{.65em}
\subcaptionbox{\label{fig:5c}}{\includegraphics[width=0.7in]{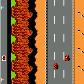}
\includegraphics[width=0.7in]{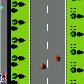}}\hspace{.65em}
\subcaptionbox{\label{fig:5d}}{\includegraphics[width=0.7in]{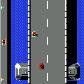}
\includegraphics[width=0.7in]{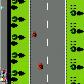}}\hspace{.65em}
\subcaptionbox{\label{fig:5e}}{\includegraphics[width=0.7in]{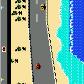}
\includegraphics[width=0.7in]{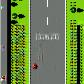}}\hspace{.65em}
\subcaptionbox{\label{fig:5f}}{\includegraphics[width=0.7in]{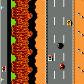}
\includegraphics[width=0.7in]{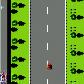}}
\caption{Left: the original frame. Right: GAN generated. The first three examples show the success cases of the GAN while the last three show representative failures: in (d) and (f) the only object generated on the road is the player's car and in (e) the diagonal shaped road of level 2 in matched to the starting point of level 1.}
\label{figure5}
\vspace{-1.1em}
\end{figure}

For the visual-transfer training, we collect $100k$ frames from each of levels 2, 3 and 4 by running an untrained agent repeatedly until we have collected sufficient samples. %(training an RL agent to reach a score of 10,000 points on Level 1 required observing above 100M frames.). 
Using the collected images we train a mapping function from each new level (target task) to the first one (source task). We use the same GAN architecture used for Breakout, but initialize the weights with Orthogonal initialization. Compared to Breakout, these tasks introduce new challenges: rather than removing a mostly static element, the GAN has to be able to change the background and road size while keeping the cars in the right position on the road. On the other hand, this setup may be closer to the one unaligned GANs are usually applied on. We restrict ourselves to collecting images from the beginning of the game, before the agent had any training. This restricts the phenomena the GAN can observe, leading to some target tasks' images without a clear corresponding situation in the first level, potentially causing unpredictable behaviors. For example, the generator matches the diagonal shaped roads to one of the first rare and unique images of level 1 (Fig. \ref{fig:5e}).

\textbf{Data Efficiency.} We measure the number of frames of game-interaction needed for the analogy-transfer method. We collect 100k frames, and then train the GAN for up to $500k$ iterations, evaluating it every $10,000$ iterations by running the game and observing the score, and pick the best scoring model. This amounts to $100k+50*F$ frames, where $F=3000$ is roughly the average number of frames in a game. This amounts to about $250k$ frames of game interaction for each transfered level, an order of magnitude fewer interaction frames than training an RL agent to achieve a comparable score.

\paragraph{Results.} The results in table \ref{table:RF} show that the visual transfer manages to achieve scores of 5350, 5350 and 2050 on level 2, 3 and 4 after $320k$, $450k$ and $270k$ GAN iterations respectively, while performing only a fraction of the game interactions required to train an RL agent from scratch to achieve these scores on these levels.

Qualitatively, the agent applies many of the abilities it gained when training on the first level, most notably driving fast, staying on the road, avoiding some cars, and, most importantly, recovering from car crashes.

\begin{table*}[!t]
    \centering
    \captionof{table}{The scores of the agent on every level of Road Fighter with and without analogy-transfer and imitation learning, as well as the number of game-interaction frames needed for the analogy transfer and for achieving a similar score with RL training, and number of frames needed for an RL agent to achieve over 10,000 points with and without imitation learning.}
    \label{table:RF}
    \begin{tabular}{l c| c c c| c c c}
        \toprule
         & \thead{Score\\ (no transfer)} 
         & \thead{Score\\ (analogy transfer)} 
         & \thead{\# Frames\\ (analogy)} 
         & \thead{\# Frames\\ (from scratch)} 
         & \thead{Score\\ (+imitation)} 
         & \thead{\#Frames\\ (imitation)}
         & \thead{\# Frames\\ (from scratch)} \\
        \midrule
        Level 2 & 0 & 5350 & 250K & 12.4M & 10230 & 38.6M & 159M \\
        Level 3 & 0 & 5350 & 250K & 31M & 10300 & 21M & 54.4M \\
 Level 4 & 0 & 2050 & 250K & 13.6M & 10460 & 13.4M & 111M\\
        \bottomrule
    \end{tabular}
    \vspace{-1em}
\end{table*}

%Our experiments presented in Table \ref{table:RF} demonstrate how an agent trained to master the first level of the game fails to follow the optimal policies on new levels, reaching $0$ points. However, with the GAN-based visual analogies the agent is able to apply some of the abilities it gained when training on the first level, most notably driving fast, staying on the road, avoiding some cars, and, most importantly, recovering from car crashes.
%The resulting agent achieves impressive scores on levels 2, 3 and 4 (5350, 5350 and 2050 points, respectively), with no additional RL training and while observing only a fraction of the frames required for training a corresponding RL agent from scratch for these levels. 

\noindent\textbf{Limitations of purely-visual transfer.} While the GAN works well in generating objects it has seen before, such as the agent's car, it does have limitations on objects it has not seen. As a result, it ends up generating differently colored cars all in red, or not generating them at all, as shown in Fig. \ref{fig:5a}, \ref{fig:5d} and \ref{fig:5f}. Colorful cars can be ``collected'' by the agent and are worth $1000$ points each. Generating them in red makes the agent avoid them, losing these extra points and achieving overall lower scores even if finishing the track. When cars are not fully generated, the agent is less likely to avoid them, and eventually crashes. 

In general, as the game progresses, the more obstacles are presented making it harder to play. On level 3, the challenge increases as the road is very narrow, making it harder for a player to avoid crashing. However, the GAN manages to generate the road in the right shape in most frames and position the cars in the matching ratio. Level 4 gets the lowest score. In this level the main obstacles are the sudden turns in the road causing the car to be very close to the sideways and the increasing dangers a player has to avoid. Theses difficulties make this level much harder than level 1 and might require more training even from a human player.

% \textbf{Zero-Shot Transfer.}\yg{This is a very good candidate for removal, can merge some of the content into qualitative results and into limitations.} The approach is successfully transferring to most tasks, achieving the best scores on levels 2 and 3. In general, as the game progresses, the more obstacles are presented making it harder to play.
% One of our best results is achieved on level 2 where the road is identical to level 1's, reaching the best score after $320k$ GAN iterations. On level 3, the challenge increases as the road is very narrow, making it harder for a player to avoid crashing. However, the GAN manages to generate the road in the right shape in most frames and position the cars in the matching ratio. Moreover, despite the challenges, due to the agent's ability to avoid crashing when colliding with cars it gets over $5000$ points after $450k$ GAN iterations. In level 4 we get the maximum after $270k$ iterations. This level is also the most challenging one to play, which might be the reason for the score being the lowest out of all tasks. The main obstacles are the sudden turns in the road causing the car to be very close to the sideways and the increasing dangers a player has to avoid. Theses difficulties make this level much harder than level 1 and might require more training even from a human player.

%\setlength\belowcaptionskip{-4ex}

\textbf{Improving with Imitation Learning.}
Using zero-shot visual transfer, the agent manages to apply 2 out of 3 of the main capabilities it gained when training on level 1. It is missing the third capability: avoiding collisions and collecting bonus cars, mainly because of bad generation. It is also challenged by winding roads (levels 3 and 4), which are not available on level 1. We now train an RL agent to play the target levels by imitating the imperfect demonstration of the visual transfer policy (see Section \ref{IL_A2C}). 

\begin{figure*}
\centering
\subcaptionbox{\label{g:lvl2}}{\includegraphics[width=1.45in]{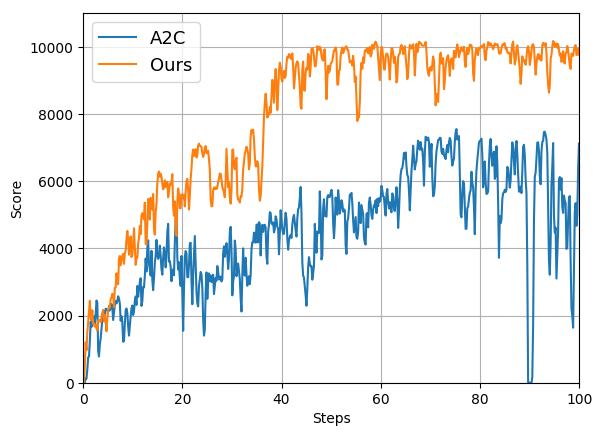}}\hspace{0.8em}
\subcaptionbox{\label{g:lvl3}}{\includegraphics[width=1.45in]{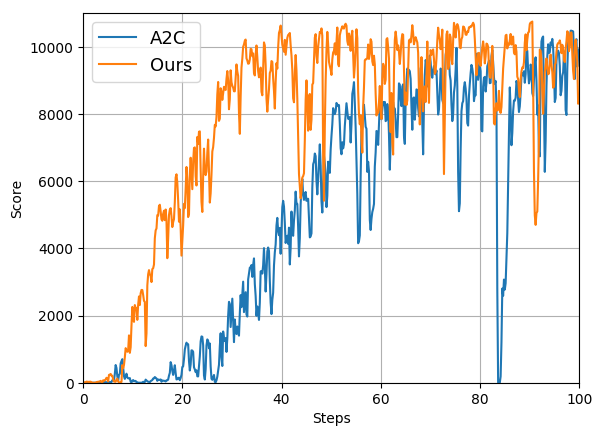}}\hspace{0.8em}
\subcaptionbox{\label{g:lvl4}}{\includegraphics[width=1.45in]{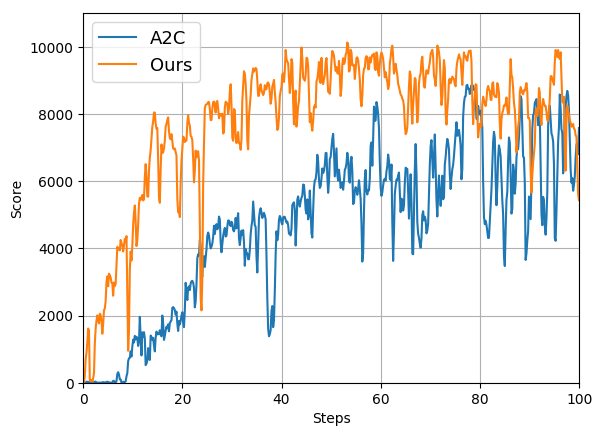}}
\caption{Learning curves of the agent trained to play levels 2 (a),3 (b) and 4 (c) of Road Fighter from scratch using A2C, and using our A2C with Imitation Learning method. Each point represents the average reward of 84 agents every 100 RL updates.}
\label{IL:GRAPHS}
\end{figure*}

The graphs in Figure \ref{IL:GRAPHS} demonstrates that the combination of visual transfer and imitation learning works well, surpassing training from scratch on the target tasks in terms of both sample efficiency \emph{and} accuracy. The final agents clear levels 2, 3 and 4 with scores of over 10,000 points on each.

%\textbf{Imitation Learning Results.} As shown in Figure \ref{IL:GRAPHS} our combination is better than training from scratch in terms of both, accuracy and sample complexity on each one of the target tasks. While the A2C agents have to explore and interact with the environment to improve the policy in each step, the imitation algorithm provide good starting behaviours to exploit and by that accelerates the process.

%Overall, the agent performs quite well using zero-shot transfer by successfully applying 2 out of 3 main capabilities it gained during training. It is missing the third capability, avoiding collisions and collecting bonus cars, mainly because of bad generation. Therefore, we further train the agent on the target tasks and use the capabilities gained for boosting the training and achieving better scores.
% We believe that these tasks and results demonstrate a success of the analogy transfer method across different levels of a video game. They also suggest a potential of performing well on additional real world tasks in which visual analogies can be made.\yg{maybe move this to the conclusions?}

\subsection{Towards Task-oriented GAN evaluation}
\label{GANEval}

Evaluating GAN models and comparing them to each other is a major challenge. Our setup introduces a natural, measurable objective for unaligned GAN evaluation: using them for visual transfer of an RL agent, and measuring the agent's performance on the translated environment.
%We propose to evaluate them by running a game in which the trained Deep RL network that learned policies based on images is now receiving images from the same domain generated with GAN, and equate a successful GAN model with one resulting in a high game score. 
We use the approach to compare Cycle-GAN and UNIT-GAN
with somewhat mixed results: UNIT works better for breakout, while the methods are mostly tied for Road Fighter, with some advantage to Cycle-GAN (exact numbers are available in the appendix, Table \ref{table:GAN}).
%and present the results in Table \ref{table:GAN}.
The main difference between the two methods is the weight-sharing constraint applied in UNIT, making the domains dependent on each other by sharing and updating the weights of one or several decoders and encoders layers. We hypothesize this constraint is an advantage in tasks where the representation of the images in the different domains are similar, such as the Breakout variants. In contrast, the more distinct Road Fighter levels could benefit from the independence in Cycle-GAN.

\section{Related Work}

Transfer Learning (TL) is a machine learning technique used to improve the training speed of a target task with knowledge learned in a source task. Pretraining and fine-tuning was proposed in \cite{Hinton504} and applied to TL in \cite{pmlr-v27-bengio12a} and \cite{pmlr-v27-mesnil12a}. In this procedure, the approach is to train the base network and then copy its first $n$ layers to the first $n$ layers of a target network. One can choose to update the feature layers transferred to the new task with the error backpropagated from its output, or they can be left frozen, meaning that they do not change during training on the new task. Unfortunately, as we have shown, while fine-tuning might have the ability to accelerate the training process is some cases, it can also have a damaging impact on others.

Generalization is a key element in training deep learning models with time or data size constraints. Recent discussions on overfitting in Deep RL algorithms \citep{DBLP:journals/corr/abs-1804-06893} encouraged better evaluation (e.g. OpenAI Retro Contest \footnote{https://contest.openai.com/}) and generalization methods.
In Atari, there are many similarities between the goals and the mechanism of the games. For this reason, there have been many works attempting to transfer between games or between different variations of the same game, one approach trying to do both is the \textit{progressive networks} \citep{DBLP:journals/corr/RusuRDSKKPH16}. A progressive network is constructed by successively training copies of A3C on each task of interest. In that work, they transferred between different games as well as from different variations of the game Pong. The drawback of this approach is the number of parameters growing quadratically with the number of tasks.
% However, even if this growth rate was improved, different tasks may require different adjustments and the predefinition of the number of layers and network representation is preventing it. 

Zero-shot generalization is a popular research topic.
%Zero-shot generalization is a discussed and researched topic nowadays.
In the context of games, \citep{DBLP:journals/corr/KanskySMELLDSPG17} performs
zero-shot transfer between modified versions of the same game using schema networks. Like us, they also demonstrate their method on the game Breakout, using Object Oriented Markov Decision Process. In contrast, we do not use the representation of the objects in the game, and wish to preserve the accomplishments of DQN and transfer using only raw data. Other attempted to achieve robust policies using learned disentangled representation of the image \citep{Higgins2017DARLAIZ}, analogies between sets of instructions \citep{DBLP:journals/corr/OhSLK17}, \textit{interactive replay} \citep{DBLP:journals/corr/abs-1711-10137} while training and learn general policies by training on multiple tasks in parallel \citep{DBLP:journals/corr/abs-1802-01561, Sohn2018MultitaskRL}.

Finally, the idea of using GANs for transfer learning and domain adaptation was explored for supervised image classification and robotics applications by several authors \citep{DBLP:journals/corr/BousmalisSDEK16, DBLP:journals/corr/abs-1711-03213, DBLP:journals/corr/0001T16, DBLP:journals/corr/abs-1709-07857}. Our work is the first to combine it with imitation learning and successfully demonstrate it in an RL setup.

Imitation learning is the paradigm of learning from demonstrations by imitating their behaviours. Combing it with RL seems natural, while RL provides more exploration of new states, imitation improves RL by providing prior knowledge to exploit. Recent works has shown a success of this combination in a difficult setting of learning from imperfect demonstrations. DQfD \cite{DBLP:journals/corr/HesterVPLSPSDOA17} and SIL \cite{DBLP:journals/corr/abs-1806-05635} merge temporal difference and imitation losses for training and prioritize better demonstrations by choosing the ones 
that are most likely to improve the current behaviour. In contrast, NAC \cite{DBLP:journals/corr/abs-1802-05313} is an RL algorithm that uses a unified actor-critic objective function that is capable of resisting poor behaviors. Our approach is similar to these works. However, we prioritize good behaviours from the start, by selecting trajectories with the highest scores. Moreover, we use a separate supervised loss function for imitation and for RL and train our agent with both as long as it benefits the learning.
% In these methods, there is supervised source domain data $(x^s_i,y^s_i)$ and unlabeled target domain data, and the GAN variant $G$ is trained to map
% source samples $x^s_i$ to target-like samples $G(x^s_i)$ . Then, a classifier is trained on the generated data $(G(x^s_i), y^s_i)$.
% Our RL-based setup is different: first, our coverage of target-domain data is very limited (we can only observe states which are reachable by the un-adapted or untrained agent). Second, we do not have access to supervised gold labels on the source domain, but only to a learned policy network. Third, interactions with the game environment provide very indirect rewards, so using this reward signal to influence the GAN training will be very inefficient. We thus opt for a different strategy: rather than mapping the source to the target domain and training on the projected signal, which is unrealistic an costly in the RL setup, we instead take a pre-trained source model and train an unaligned GAN to map from the target domain back to the source domain, in order to re-use the source model's knowledge and apply it to the target domain data. We believe this form of usage of GAN for transfer learning is novel.

\section{Conclusions}

We demonstrated the lack of generalization by looking at artificially constructed visual variants of a game (Breakout), and different levels of a game (Road Fighter). We further show that transfer learning by fine-tuning fails. The policies learned using model-free RL algorithms on the original game are not directly transferred to the modified games even when the changes are irrelevant to the game's dynamics.

We present a new approach for transfer learning between related RL environments using GANs without the need for any additional training of the RL agent, and while requiring orders of magnitude less interactions with the environment. We further suggest this setup as a way to evaluate GAN architectures by observing their behavior on concrete tasks, revealing differences between the Cycle-GAN and UNIT-GAN architectures.  
While we report a success in analogy transfer using Unaligned GANs, we also encountered limitations in the generation process that made it difficult for the agent to maximize the results on the Road Fighter's tasks. We overcome these difficulties by using the imperfect behaviours as demonstrations to improve and accelerate the RL training on each one of the target tasks.
We believe that these tasks and results demonstrate a success of the analogy transfer method across different levels of a video game. They also suggest a potential of performing well on additional real world tasks in which visual analogies can be made.
%We believe our approach is applicable to cases involving both direct and less direct mapping between environments, as long as an image-to-image translation exist. 
%In future work, we plan to explore a tighter integration between the analogy transfer method and the RL training process, to facilitate better performance where dynamic adjustments are needed in addition to the visual mapping.


\section*{Acknowledgements}
We thank Hal Daumé III for the helpful discussions on the Imitation Learning algorithm during the development of the work. 

\bibliography{example_paper}

\begin{thebibliography}{35}
\providecommand{\natexlab}[1]{#1}
\providecommand{\url}[1]{\texttt{#1}}
\expandafter\ifx\csname urlstyle\endcsname\relax
  \providecommand{\doi}[1]{doi: #1}\else
  \providecommand{\doi}{doi: \begingroup \urlstyle{rm}\Url}\fi

\bibitem[Bengio(2012)]{pmlr-v27-bengio12a}
Bengio, Y.
\newblock Deep learning of representations for unsupervised and transfer
  learning.
\newblock In Guyon, I., Dror, G., Lemaire, V., Taylor, G., and Silver, D.
  (eds.), \emph{Proceedings of ICML Workshop on Unsupervised and Transfer
  Learning}, volume~27 of \emph{Proceedings of Machine Learning Research}, pp.\
   17--36, Bellevue, Washington, USA, 02 Jul 2012. PMLR.
\newblock URL \url{http://proceedings.mlr.press/v27/bengio12a.html}.

\bibitem[Bousmalis et~al.(2016)Bousmalis, Silberman, Dohan, Erhan, and
  Krishnan]{DBLP:journals/corr/BousmalisSDEK16}
Bousmalis, K., Silberman, N., Dohan, D., Erhan, D., and Krishnan, D.
\newblock Unsupervised pixel-level domain adaptation with generative
  adversarial networks.
\newblock \emph{CoRR}, abs/1612.05424, 2016.
\newblock URL \url{http://arxiv.org/abs/1612.05424}.

\bibitem[Bousmalis et~al.(2017)Bousmalis, Irpan, Wohlhart, Bai, Kelcey,
  Kalakrishnan, Downs, Ibarz, Pastor, Konolige, Levine, and
  Vanhoucke]{DBLP:journals/corr/abs-1709-07857}
Bousmalis, K., Irpan, A., Wohlhart, P., Bai, Y., Kelcey, M., Kalakrishnan, M.,
  Downs, L., Ibarz, J., Pastor, P., Konolige, K., Levine, S., and Vanhoucke, V.
\newblock Using simulation and domain adaptation to improve efficiency of deep
  robotic grasping.
\newblock \emph{CoRR}, abs/1709.07857, 2017.
\newblock URL \url{http://arxiv.org/abs/1709.07857}.

\bibitem[Bruce et~al.(2017)Bruce, S{\"{u}}nderhauf, Mirowski, Hadsell, and
  Milford]{DBLP:journals/corr/abs-1711-10137}
Bruce, J., S{\"{u}}nderhauf, N., Mirowski, P., Hadsell, R., and Milford, M.
\newblock One-shot reinforcement learning for robot navigation with interactive
  replay.
\newblock \emph{CoRR}, abs/1711.10137, 2017.
\newblock URL \url{http://arxiv.org/abs/1711.10137}.

\bibitem[Dauphin et~al.(2012)Dauphin, Glorot, Rifai, Bengio, Goodfellow,
  Lavoie, Muller, Desjardins, Warde-Farley, Vincent, Courville, and
  Bergstra]{pmlr-v27-mesnil12a}
Dauphin, G. M.~Y., Glorot, X., Rifai, S., Bengio, Y., Goodfellow, I., Lavoie,
  E., Muller, X., Desjardins, G., Warde-Farley, D., Vincent, P., Courville, A.,
  and Bergstra, J.
\newblock Unsupervised and transfer learning challenge: a deep learning
  approach.
\newblock In Guyon, I., Dror, G., Lemaire, V., Taylor, G., and Silver, D.
  (eds.), \emph{Proceedings of ICML Workshop on Unsupervised and Transfer
  Learning}, volume~27 of \emph{Proceedings of Machine Learning Research}, pp.\
   97--110, Bellevue, Washington, USA, 02 Jul 2012. PMLR.
\newblock URL \url{http://proceedings.mlr.press/v27/mesnil12a.html}.

\bibitem[Espeholt et~al.(2018)Espeholt, Soyer, Munos, Simonyan, Mnih, Ward,
  Doron, Firoiu, Harley, Dunning, Legg, and
  Kavukcuoglu]{DBLP:journals/corr/abs-1802-01561}
Espeholt, L., Soyer, H., Munos, R., Simonyan, K., Mnih, V., Ward, T., Doron,
  Y., Firoiu, V., Harley, T., Dunning, I., Legg, S., and Kavukcuoglu, K.
\newblock {IMPALA:} scalable distributed deep-rl with importance weighted
  actor-learner architectures.
\newblock \emph{CoRR}, abs/1802.01561, 2018.
\newblock URL \url{http://arxiv.org/abs/1802.01561}.

\bibitem[Fernando et~al.(2017)Fernando, Banarse, Blundell, Zwols, Ha, Rusu,
  Pritzel, and Wierstra]{DBLP:journals/corr/FernandoBBZHRPW17}
Fernando, C., Banarse, D., Blundell, C., Zwols, Y., Ha, D., Rusu, A.~A.,
  Pritzel, A., and Wierstra, D.
\newblock Pathnet: Evolution channels gradient descent in super neural
  networks.
\newblock \emph{CoRR}, abs/1701.08734, 2017.
\newblock URL \url{http://arxiv.org/abs/1701.08734}.

\bibitem[Gao et~al.(2018)Gao, Xu, Lin, Yu, Levine, and
  Darrell]{DBLP:journals/corr/abs-1802-05313}
Gao, Y., Xu, H., Lin, J., Yu, F., Levine, S., and Darrell, T.
\newblock Reinforcement learning from imperfect demonstrations.
\newblock \emph{CoRR}, abs/1802.05313, 2018.
\newblock URL \url{http://arxiv.org/abs/1802.05313}.

\bibitem[Glorot \& Bengio(2010)Glorot and Bengio]{Glorot10understandingthe}
Glorot, X. and Bengio, Y.
\newblock Understanding the difficulty of training deep feedforward neural
  networks.
\newblock In \emph{In Proceedings of the International Conference on Artificial
  Intelligence and Statistics (AISTATS’10). Society for Artificial
  Intelligence and Statistics}, 2010.

\bibitem[Goodfellow et~al.(2014)Goodfellow, Pouget-Abadie, Mirza, Xu,
  Warde-Farley, Ozair, Courville, and Bengio]{NIPS2014_5423}
Goodfellow, I., Pouget-Abadie, J., Mirza, M., Xu, B., Warde-Farley, D., Ozair,
  S., Courville, A., and Bengio, Y.
\newblock Generative adversarial nets.
\newblock In Ghahramani, Z., Welling, M., Cortes, C., Lawrence, N.~D., and
  Weinberger, K.~Q. (eds.), \emph{Advances in Neural Information Processing
  Systems 27}, pp.\  2672--2680. Curran Associates, Inc., 2014.
\newblock URL
  \url{http://papers.nips.cc/paper/5423-generative-adversarial-nets.pdf}.

\bibitem[Hester et~al.(2017)Hester, Vecerik, Pietquin, Lanctot, Schaul, Piot,
  Sendonaris, Dulac{-}Arnold, Osband, Agapiou, Leibo, and
  Gruslys]{DBLP:journals/corr/HesterVPLSPSDOA17}
Hester, T., Vecerik, M., Pietquin, O., Lanctot, M., Schaul, T., Piot, B.,
  Sendonaris, A., Dulac{-}Arnold, G., Osband, I., Agapiou, J., Leibo, J.~Z.,
  and Gruslys, A.
\newblock Learning from demonstrations for real world reinforcement learning.
\newblock \emph{CoRR}, abs/1704.03732, 2017.
\newblock URL \url{http://arxiv.org/abs/1704.03732}.

\bibitem[Higgins et~al.(2017)Higgins, Pal, Rusu, Matthey, Burgess, Pritzel,
  Botvinick, Blundell, and Lerchner]{Higgins2017DARLAIZ}
Higgins, I., Pal, A., Rusu, A.~A., Matthey, L., Burgess, C., Pritzel, A.,
  Botvinick, M.~M., Blundell, C., and Lerchner, A.
\newblock Darla: Improving zero-shot transfer in reinforcement learning.
\newblock In \emph{ICML}, 2017.

\bibitem[Hinton \& Salakhutdinov(2006)Hinton and Salakhutdinov]{Hinton504}
Hinton, G.~E. and Salakhutdinov, R.~R.
\newblock Reducing the dimensionality of data with neural networks.
\newblock \emph{Science}, 313\penalty0 (5786):\penalty0 504--507, 2006.
\newblock ISSN 0036-8075.
\newblock \doi{10.1126/science.1127647}.
\newblock URL \url{http://science.sciencemag.org/content/313/5786/504}.

\bibitem[Ho \& Ermon(2016)Ho and Ermon]{DBLP:journals/corr/HoE16}
Ho, J. and Ermon, S.
\newblock Generative adversarial imitation learning.
\newblock \emph{CoRR}, abs/1606.03476, 2016.
\newblock URL \url{http://arxiv.org/abs/1606.03476}.

\bibitem[Hoffman et~al.(2017)Hoffman, Tzeng, Park, Zhu, Isola, Saenko, Efros,
  and Darrell]{DBLP:journals/corr/abs-1711-03213}
Hoffman, J., Tzeng, E., Park, T., Zhu, J., Isola, P., Saenko, K., Efros, A.~A.,
  and Darrell, T.
\newblock Cycada: Cycle-consistent adversarial domain adaptation.
\newblock \emph{CoRR}, abs/1711.03213, 2017.
\newblock URL \url{http://arxiv.org/abs/1711.03213}.

\bibitem[Isola et~al.(2016)Isola, Zhu, Zhou, and
  Efros]{DBLP:journals/corr/IsolaZZE16}
Isola, P., Zhu, J., Zhou, T., and Efros, A.~A.
\newblock Image-to-image translation with conditional adversarial networks.
\newblock \emph{CoRR}, abs/1611.07004, 2016.
\newblock URL \url{http://arxiv.org/abs/1611.07004}.

\bibitem[Kansky et~al.(2017)Kansky, Silver, M{\'{e}}ly, Eldawy,
  L{\'{a}}zaro{-}Gredilla, Lou, Dorfman, Sidor, Phoenix, and
  George]{DBLP:journals/corr/KanskySMELLDSPG17}
Kansky, K., Silver, T., M{\'{e}}ly, D.~A., Eldawy, M., L{\'{a}}zaro{-}Gredilla,
  M., Lou, X., Dorfman, N., Sidor, S., Phoenix, D.~S., and George, D.
\newblock Schema networks: Zero-shot transfer with a generative causal model of
  intuitive physics.
\newblock \emph{CoRR}, abs/1706.04317, 2017.
\newblock URL \url{http://arxiv.org/abs/1706.04317}.

\bibitem[Kim et~al.(2017)Kim, Cha, Kim, Lee, and
  Kim]{DBLP:journals/corr/KimCKLK17}
Kim, T., Cha, M., Kim, H., Lee, J.~K., and Kim, J.
\newblock Learning to discover cross-domain relations with generative
  adversarial networks.
\newblock \emph{CoRR}, abs/1703.05192, 2017.
\newblock URL \url{http://arxiv.org/abs/1703.05192}.

\bibitem[Kostrikov(2018{\natexlab{a}})]{pytorchaaac}
Kostrikov, I.
\newblock Pytorch implementations of asynchronous advantage actor critic.
\newblock \url{https://github.com/ikostrikov/pytorch-a3c}, 2018{\natexlab{a}}.

\bibitem[Kostrikov(2018{\natexlab{b}})]{pytorchrl}
Kostrikov, I.
\newblock Pytorch implementations of reinforcement learning algorithms.
\newblock \url{https://github.com/ikostrikov/pytorch-a2c-ppo-acktr},
  2018{\natexlab{b}}.

\bibitem[Liu \& Tuzel(2016)Liu and Tuzel]{DBLP:journals/corr/0001T16}
Liu, M. and Tuzel, O.
\newblock Coupled generative adversarial networks.
\newblock \emph{CoRR}, abs/1606.07536, 2016.
\newblock URL \url{http://arxiv.org/abs/1606.07536}.

\bibitem[Liu et~al.(2017)Liu, Breuel, and Kautz]{DBLP:journals/corr/LiuBK17}
Liu, M., Breuel, T., and Kautz, J.
\newblock Unsupervised image-to-image translation networks.
\newblock \emph{CoRR}, abs/1703.00848, 2017.
\newblock URL \url{http://arxiv.org/abs/1703.00848}.

\bibitem[Mnih et~al.(2013)Mnih, Kavukcuoglu, Silver, Graves, Antonoglou,
  Wierstra, and Riedmillera]{DBLP:journals/corr/MnihKSGAWR13}
Mnih, V., Kavukcuoglu, K., Silver, D., Graves, A., Antonoglou, I., Wierstra,
  D., and Riedmillera, M.~A.
\newblock Playing atari with deep reinforcement learning.
\newblock \emph{CoRR}, abs/1312.5602, 2013.
\newblock URL \url{http://arxiv.org/abs/1312.5602}.

\bibitem[Mnih et~al.(2016)Mnih, Badia, Mirza, Graves, Lillicrap, Harley,
  Silver, and Kavukcuoglu]{DBLP:journals/corr/MnihBMGLHSK16}
Mnih, V., Badia, A.~P., Mirza, M., Graves, A., Lillicrap, T.~P., Harley, T.,
  Silver, D., and Kavukcuoglu, K.
\newblock Asynchronous methods for deep reinforcement learning.
\newblock \emph{CoRR}, abs/1602.01783, 2016.
\newblock URL \url{http://arxiv.org/abs/1602.01783}.

\bibitem[Oh et~al.(2017)Oh, Singh, Lee, and Kohli]{DBLP:journals/corr/OhSLK17}
Oh, J., Singh, S.~P., Lee, H., and Kohli, P.
\newblock Zero-shot task generalization with multi-task deep reinforcement
  learning.
\newblock \emph{CoRR}, abs/1706.05064, 2017.
\newblock URL \url{http://arxiv.org/abs/1706.05064}.

\bibitem[Oh et~al.(2018)Oh, Guo, Singh, and
  Lee]{DBLP:journals/corr/abs-1806-05635}
Oh, J., Guo, Y., Singh, S., and Lee, H.
\newblock Self-imitation learning.
\newblock \emph{CoRR}, abs/1806.05635, 2018.
\newblock URL \url{http://arxiv.org/abs/1806.05635}.

\bibitem[Ross et~al.(2010)Ross, Gordon, and
  Bagnell]{DBLP:journals/corr/abs-1011-0686}
Ross, S., Gordon, G.~J., and Bagnell, J.~A.
\newblock No-regret reductions for imitation learning and structured
  prediction.
\newblock \emph{CoRR}, abs/1011.0686, 2010.
\newblock URL \url{http://arxiv.org/abs/1011.0686}.

\bibitem[Rusu et~al.(2016)Rusu, Rabinowitz, Desjardins, Soyer, Kirkpatrick,
  Kavukcuoglu, Pascanu, and Hadsell]{DBLP:journals/corr/RusuRDSKKPH16}
Rusu, A.~A., Rabinowitz, N.~C., Desjardins, G., Soyer, H., Kirkpatrick, J.,
  Kavukcuoglu, K., Pascanu, R., and Hadsell, R.
\newblock Progressive neural networks.
\newblock \emph{CoRR}, abs/1606.04671, 2016.
\newblock URL \url{http://arxiv.org/abs/1606.04671}.

\bibitem[Schaul et~al.(2015)Schaul, Quan, Antonoglou, and
  Silver]{DBLP:journals/corr/SchaulQAS15}
Schaul, T., Quan, J., Antonoglou, I., and Silver, D.
\newblock Prioritized experience replay.
\newblock \emph{CoRR}, abs/1511.05952, 2015.
\newblock URL \url{http://arxiv.org/abs/1511.05952}.

\bibitem[Sohn et~al.(2018)Sohn, Oh, and Lee]{Sohn2018MultitaskRL}
Sohn, S., Oh, J., and Lee, H.
\newblock Multitask reinforcement learning for zero-shot generalization with
  subtask dependencies.
\newblock \emph{CoRR}, abs/1807.07665, 2018.

\bibitem[Torabi et~al.(2018)Torabi, Warnell, and
  Stone]{DBLP:journals/corr/abs-1805-01954}
Torabi, F., Warnell, G., and Stone, P.
\newblock Behavioral cloning from observation.
\newblock \emph{CoRR}, abs/1805.01954, 2018.
\newblock URL \url{http://arxiv.org/abs/1805.01954}.

\bibitem[Yi et~al.(2017)Yi, Zhang, Tan, and Gong]{DBLP:journals/corr/YiZTG17}
Yi, Z., Zhang, H., Tan, P., and Gong, M.
\newblock Dualgan: Unsupervised dual learning for image-to-image translation.
\newblock \emph{CoRR}, abs/1704.02510, 2017.
\newblock URL \url{http://arxiv.org/abs/1704.02510}.

\bibitem[Yosinski et~al.(2014)Yosinski, Clune, Bengio, and
  Lipson]{Yosinski:2014:TFD:2969033.2969197}
Yosinski, J., Clune, J., Bengio, Y., and Lipson, H.
\newblock How transferable are features in deep neural networks?
\newblock In \emph{Proceedings of the 27th International Conference on Neural
  Information Processing Systems - Volume 2}, NIPS'14, pp.\  3320--3328,
  Cambridge, MA, USA, 2014. MIT Press.
\newblock URL \url{http://dl.acm.org/citation.cfm?id=2969033.2969197}.

\bibitem[Zhang et~al.(2018)Zhang, Vinyals, Munos, and
  Bengio]{DBLP:journals/corr/abs-1804-06893}
Zhang, C., Vinyals, O., Munos, R., and Bengio, S.
\newblock A study on overfitting in deep reinforcement learning.
\newblock \emph{CoRR}, abs/1804.06893, 2018.
\newblock URL \url{http://arxiv.org/abs/1804.06893}.

\bibitem[Zhu et~al.(2017)Zhu, Park, Isola, and
  Efros]{DBLP:journals/corr/ZhuPIE17}
Zhu, J., Park, T., Isola, P., and Efros, A.~A.
\newblock Unpaired image-to-image translation using cycle-consistent
  adversarial networks.
\newblock \emph{CoRR}, abs/1703.10593, 2017.
\newblock URL \url{http://arxiv.org/abs/1703.10593}.

\end{thebibliography}
\bibliographystyle{icml2019}

% \appendix
% \section{Do \emph{not} have an appendix here}

% \textbf{\emph{Do not put content after the references.}}
% %
% Put anything that you might normally include after the references in a separate
% supplementary file.

% We recommend that you build supplementary material in a separate document.
% If you must create one PDF and cut it up, please be careful to use a tool that
% doesn't alter the margins, and that doesn't aggressively rewrite the PDF file.
% pdftk usually works fine. 

% \textbf{Please do not use Apple's preview to cut off supplementary material.} In
% previous years it has altered margins, and created headaches at the camera-ready
% stage. 
%%%%%%%%%%%%%%%%%%%%%%%%%%%%%%%%%%%%%%%%%%%%%%%%%%%%%%%%%%%%%%%%%%%%%%%%%%%%%%%
%%%%%%%%%%%%%%%%%%%%%%%%%%%%%%%%%%%%%%%%%%%%%%%%%%%%%%%%%%%%%%%%%%%%%%%%%%%%%%%

\clearpage
\onecolumn
\appendix
\pagenumbering{gobble}
\title{Supplementary Material}
\date{}
\maketitle 
\vspace{-6em}
\section{Experimental Setup}

\subsection{A3C}
\label{A3C-setup}
For our Breakout experiments we use the standard high-performance architecture implemented in \citep{pytorchaaac}. 
%$4$ convolutional layers with $32$ $3x3$ filters and a stride length of $2$, connected to an LSTM layer with $256$ cells whose output is linearly fed into two output layers, one produces the next optimal action (actor) and the other, the approximated value (critic).

\begin{table}[!h]
    \centering
    \captionof{table}{A3C hyperparameters}
    \label{table-p:A3C}
    \begin{tabular}{l c}
        \toprule
         Hyperparameter & Value \\
        \midrule
        architecture & LSTM-A3C \\
        state size & $1 \times 80 \times 80$\\
        \# actor learners & $32$ \\
        discount rate & $0.99$ \\
        Adam learning rate & $0.0001$ \\
        step-returns & $20$ \\
        entropy regularization weight & $0.01$ \\
        \bottomrule
    \end{tabular}
\end{table}

\subsection{A2C}
We use the implementation in \citep{pytorchrl} for comparison and as a skeleton for our method implementation.

\begin{table}[!h]
    \centering
    \captionof{table}{A2C hyperparameters}
    \label{table-p:A2C}
    \begin{tabular}{l c}
        \toprule
         Hyperparameter & Value \\
        \midrule
        architecture & FF-A2C \\
        state size & $4 \times 84 \times 84$\\
        \# actor learners & $84$ \\
        discount rate & $0.99$ \\
        RMSprop learning rate & $0.0007$ \\
        step-returns & $20$ \\
        entropy regularization weight & $0.01$ \\
        \bottomrule
    \end{tabular}
\end{table}

\subsection{A2C with Imitation Learning}

\begin{table}[!h]
    \centering
    \captionof{table}{A2C with Imitation Learning algorithm hyperparameters}
    \label{table-p:IL}
    \begin{tabular}{l c}
        \toprule
         Hyperparameter & Value \\
        \midrule
        \textit{trajectories} & $5$\\
        $\beta_1$ & $0.75$ \\
        $\beta_2$ & $0.6$ \\
        \textit{Supervised\_Iterations} & $500$ \\
        SGD learning rate & $0.0007$ \\
        SGD momentum & $0.9$ \\
        $b$ & $4$\\
        \textit{op\_interval} & $100$ \\
        \bottomrule
    \end{tabular}
\end{table}

\clearpage

\section{Fine-tuning Settings}
\label{ft-settings}

We consider the following settings for our Fine-tuning experiments on Breakout:
\begin{itemize}
  \item From-Scratch: The game is being trained from scratch on the target game.
  \item Full-FT: All of the layers are initialized with the weights of the source task and are fine-tuned on the target task.
  \item Random-Output: The convolutional layers and the LSTM layer are initialized with the weights of the source task and are fine-tuned on the target task. The output layers are initialized randomly.
  \item Partial-FT: All of the layers are initialized with the weights of the source task. The three first convolutional layers are kept frozen, and the rest are fine-tuned on the target task.
\item Partial-Random-FT: The three first convolutional layers are initialized with the weights of the source task and are kept frozen, and the rest are initialized randomly.
\end{itemize}

\section{GAN Comparison Evaluation}

\begin{table}[!h]
\centering
\caption{The scores accumulated by an Actor-Critic RL agent using UNIT and Cycle-GAN. We examine both methods by running the RL agent with each every $1000$ GAN training iterations and considering the maximum score after $500k$ iterations.}
 \label{table:GAN}
  \begin{tabular}{@{} l
                @{\hspace*{\lengtha}}     S
                @{\hspace*{\lengtha}}S
                @{\hspace*{\lengtha}}S
                @{\hspace*{\lengtha}}S @{}}
    \toprule
    \multirow{2}{*}{Method} &
      \multicolumn{2}{c}{\bfseries UNIT} &
      \multicolumn{2}{c}{\bfseries CycleGAN} \\
      \cmidrule(l){2-3} \cmidrule(l){4-5}
      & {Frames} & {Score} & {Frames} & {Score}  \\
      \midrule
A Constant Rectangle & {333K} & 399 & {358K} & 26 \\
 A Moving Square & {384K} & 300 & {338K} & 360 \\
 Green Lines & {378K} & 314 & {172K} & 273 \\
 Diagonals & {380K} & 338 & {239K} & 253 \\
  \midrule
 Road Fighter - Level 2 & {274K} & 5750 & {51K} & 6000 \\
 Road Fighter - Level 3 & {450K} & 5350 & {20K} & 3200 \\
 Road Fighter - Level 4 & {176K} & 2300 & {102K} & 2700 \\
    \bottomrule
  \end{tabular}
\end{table}

\end{document}